%% file: paper.tex
\documentclass[singlecolumn]{cuapreprint}
\usepackage{amsmath,amsfonts,amssymb}
\usepackage{float,array,tabularx,colortbl,tikz}
\usepackage{url}
\definecolor{ResultGroup}{HTML}{E6DCF2}
\definecolor{ResultOurs}{HTML}{F1EBF7}
\definecolor{AblFull}{HTML}{7851B8}
\definecolor{AblRemoved}{HTML}{8B8B8B}
\newcommand{\thead}[1]{\textbf{#1}}
\newcommand{\lead}[1]{\textbf{#1}}
\newcommand{\tablefont}{\small\setlength{\tabcolsep}{4pt}\renewcommand{\arraystretch}{1.10}}

\definecolor{CaseInk}{HTML}{67448D}
\definecolor{CaseLine}{HTML}{BCA8D0}
\definecolor{CaseGray}{HTML}{F6F5F7}
\newcommand{\casebox}[3]{%
  \par\noindent\begingroup
  \setlength{\fboxsep}{0pt}\setlength{\fboxrule}{0.5pt}%
  \fcolorbox{CaseLine}{white}{\begin{minipage}{\dimexpr\linewidth-1pt\relax}
    \colorbox{#1}{\parbox{\linewidth}{\vspace{4pt}\hspace{8pt}\textbf{#2}\vspace{4pt}}}%
    \par\vspace{5pt}\noindent\hspace{8pt}\begin{minipage}{\dimexpr\linewidth-16pt\relax}
      \raggedright #3\par
    \end{minipage}\par\vspace{6pt}
  \end{minipage}}\par\endgroup\vspace{6pt}}
\newcommand{\casefield}[2]{\textbf{#1}\ #2\par}

\titlespacing*{\section}{0pt}{13pt plus 2pt}{7pt}
\titlespacing*{\subsection}{0pt}{10pt plus 2pt}{5pt}
\title{CUA-Sandbox: Efficient Environments for\\Computer-Use Agent Reinforcement Learning}
\author[1,3,*]{Xin Yan}
\author[2,*]{Zhengbo Jiao}
\author[3]{Jiaqi Liu}
\author[4]{Zhenglin Wan}
\author[5]{SiYuan Ma}
\author[6]{Xuliang Yu}
\author[7]{Tianyi Jiang}
\author[5]{Chubin Zhang}
\author[4]{Pengfei Zhou}
\author[2]{Wangbo Zhao}
\author[1,\dagger]{Xingrui Yu}
\author[5]{Bo An}
\author[4]{Yang You}
\author[1]{Ivor Tsang}

\affiliation[1]{A*STAR}
\affiliation[2]{HKUST}
\affiliation[3]{BNU}
\affiliation[4]{NUS}
\affiliation[5]{NTU}
\affiliation[6]{ZJU}
\affiliation[7]{PKU}
\contribution[*]{Equal contribution}
\contribution[\dagger]{Corresponding author}
\correspondence{\email{frostlin2005@gmail.com};\ \email{yu\_xingrui@a-star.edu.sg}}
\metadata[Github Link]{\href{https://github.com/QQQQQQ11234/CUA-Sandbox-Efficient-Environments-for-Computer-Use-Reinforcement-Learning}{\includegraphics[height=0.32cm]{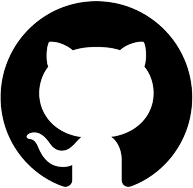}}}
\abstract{
Reinforcement learning enables computer-use agents to improve through interaction with real software environments, including websites and desktop applications. However, conventional deployments replicate an initialized runtime for each independent rollout, even when trajectories use the same software, incurring repeated memory and initialization costs as the number of parallel environments grows. Does an independent computer-use environment require an independent execution runtime? Our key observation is that trajectories require independent mutable state, while initialized application runtimes can be reused across concurrently evolving environments, making state the natural unit of environment independence. Guided by this observation, we introduce CUA-Sandbox, which separates private state capsules from shared runtimes through state-scoped execution and transactional lifecycle operations, including resets and branches, while retaining the original software interfaces and task evaluators. Experiments show comparable or improved task success relative to Docker, while substantially reducing rollout and resource costs. CUA-Sandbox achieves up to a 6.20$\times$ increase in rollout throughput, a 9.2$\times$ reduction in per-environment memory, and a 504$\times$ reduction in incremental storage.
}
\begin{document}
\maketitle

\begin{figure}[H]
\centering
\includegraphics[width=0.82\linewidth]{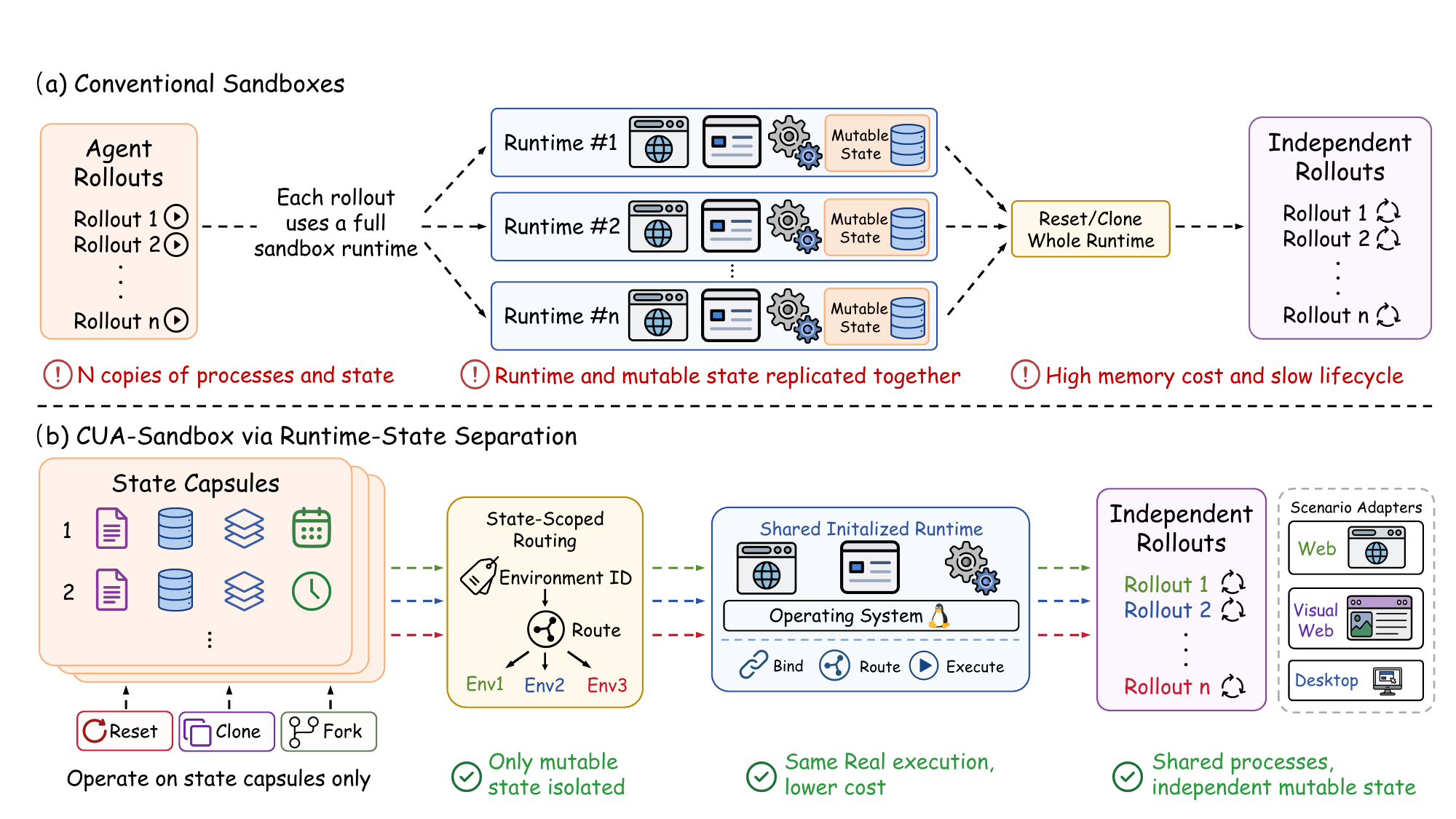}
\caption{\lead{Environment independence through runtime--state separation.} (a) Conventional sandboxes allocate a complete initialized runtime and mutable state to each rollout, repeating process memory and lifecycle work. (b) CUA-Sandbox shares initialized processes while storing each environment's mutable resources in a private state capsule. State-scoped bindings route actions to the correct capsule, allowing trajectories to evolve independently. Reset, clone, and fork operate on capsule state instead of rebuilding the runtime. Scenario adapters retain the original web, visual-web, and desktop interaction interfaces, so runtime reuse preserves execution in real software.}
\label{fig:runtime-state}
\end{figure}
\clearpage

\section{Introduction}
Computer-use agents (CUAs) operate websites, applications, and operating systems through the same interfaces available to people. Their scope has expanded from web tasks~\citep{zhou2024webarena,koh2024visualwebarena} to desktop and mobile systems~\citep{xie2024osworld,rawles2025androidworld}, including long-horizon workflows that span multiple applications~\citep{yuan2026osworld2,wang2025opencua}. Learning these behaviors increasingly relies on reinforcement learning (RL), where agents improve through repeated interaction with executable software environments~\citep{agentgym,cuagym}. Each rollout changes files, sessions, databases, or application state, so parallel trajectories require independent mutable environments.

Providing this independence is expensive. Conventional systems typically instantiate a complete container, virtual machine, or application stack for each environment. Rollouts repeatedly initialize the same software even when they differ only in the state produced by their actions. Memory and lifecycle costs therefore grow with the number of environments, limiting the experience that a training system can collect under a fixed resource budget. This coupling raises a basic question: \emph{does an independent computer-use environment require an independent execution runtime?}

Existing approaches address different parts of this cost. Learned world and experience models approximate transitions~\citep{agentworldmodel,webworld,dreamgym}, while code-generated environments expand the supply of executable tasks~\citep{scaleenv,agentworld,guigenesis}. These approaches can provide cheaper experience, but their fidelity is bounded by the dynamics and dependencies they capture. Real software introduces persistent sessions, hidden service state, rendering semantics, and distributed side effects that matter over long trajectories. Containers and microVMs preserve real execution~\citep{merkel2014docker,firecracker}, and recent systems reduce provisioning and checkpointing costs~\citep{webserv,sweminisandbox,crab}. Yet a separate initialized runtime commonly remains the unit of environment independence.

Our observation is that application code and initialized services can remain common across rollouts, while their mutable state must diverge. Figure~\ref{fig:runtime-state} illustrates this distinction. We develop \mbox{\textbf{CUA-Sandbox}}, which maintains a private state capsule for each logical environment and binds execution to the appropriate capsule. The system coordinates state transitions with ongoing requests, allowing reset and branching without rebuilding the complete runtime. Task-relevant writes remain private, and resources that cannot be safely shared use private backends.

We evaluate CUA-Sandbox on WebArena, VisualWebArena, and OSWorld through paired inference, matched agent post-training, and systems measurements. The inference study tests whether changing the execution backend preserves task success, while the post-training study examines whether agents retain the benefits of supervised fine-tuning and on-policy reinforcement learning. Rollout and operation-level measurements quantify throughput, resource use, and lifecycle costs. Component ablations separate the savings from copy-on-write, the overhead of managing non-database state, and the correctness effects of transactional publication. Together, these experiments show that runtime reuse reduces environment costs while maintaining measured agent performance.
\begin{itemize}
\item \textbf{Runtime--state separation.} We separate environment-specific state and bindings from reusable application runtimes.
\item \textbf{CUA-Sandbox for runtime reuse.} We introduce state capsules, state-scoped execution, and transactional lifecycles across web and desktop environments while retaining their interaction and evaluation interfaces.
\item \textbf{Efficiency and task fidelity.} Paired inference and post-training evaluations support task fidelity across web and desktop environments. Experiments show comparable or improved task success relative to Docker,
while substantially reducing rollout and resource costs. CUA-Sandbox
achieves up to a 6.20$\times$ increase in rollout throughput, a 9.2$\times$
reduction in per-environment memory, and a 504$\times$ reduction in
incremental storage.
\end{itemize}
\section{Related Work}
\begin{figure}[t]
    \centering
    \includegraphics[width=\linewidth]{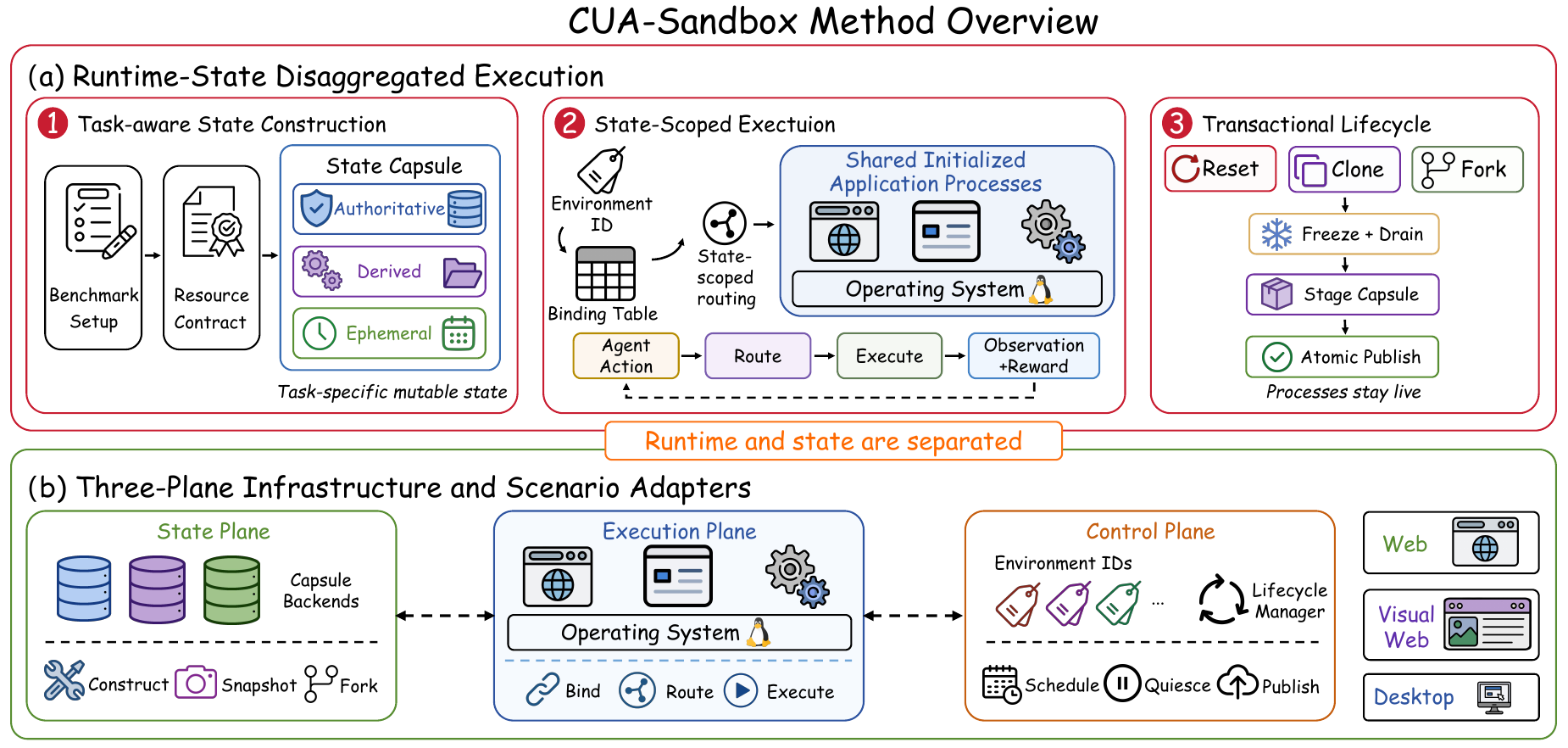}
    \caption{\lead{CUA-Sandbox architecture and execution flow.} (a) Task-aware construction turns benchmark setup and resource contracts into capsules of authoritative, derived, and ephemeral state. Environment IDs and binding tables route actions through shared initialized processes to produce observations and rewards. Reset, clone, and fork freeze and drain ongoing work, stage successor capsules, and atomically publish their bindings while processes remain live. (b) The state plane manages capsule backends; the execution plane binds and executes actions; the control plane schedules environments and coordinates lifecycle transitions. Scenario adapters expose this infrastructure to web, visual-web, and desktop tasks.}
    \label{fig:cua-overview}
\end{figure}

\textbf{Computer-Use Agents.} Early web-based computer use introduced interactive environments and demonstrations. WebShop~\citep{yao2022webshop} framed product search and purchase as interaction, while Mind2Web~\citep{deng2023mind2web} collected trajectories across diverse websites. WebArena~\citep{zhou2024webarena} established self-hosted, stateful websites with functional evaluation; VisualWebArena~\citep{koh2024visualwebarena} added visually grounded tasks, and WorkArena~\citep{drouin2024workarena} extended evaluation to enterprise workflows. WebVoyager~\citep{he2024webvoyager} studied interaction on live websites, while SeeClick~\citep{cheng2024seeclick} advanced visual grounding across interfaces. Beyond browsers, OSWorld~\citep{xie2024osworld} and Windows Agent Arena~\citep{bonatti2024windowsagentarena} covered desktop systems, AndroidWorld~\citep{rawles2025androidworld} targeted mobile devices, and OSWorld 2.0~\citep{yuan2026osworld2} introduced long-horizon professional workflows. Recent general computer-use agents include UI-TARS~\citep{qin2025uitars} and OpenCUA~\citep{wang2025opencua}, which scale data and model training, followed by Qwen-UI-Agent~\citep{zhou2026qwenuiagent} and Qwen-CUA~\citep{lu2026qwencua} across web, desktop, and mobile platforms. However, broader platforms, longer horizons, and richer state require many complete environment instances, making execution infrastructure a growing bottleneck for scalable training and evaluation.

\textbf{Reinforcement Learning Environments.} Early agentic RL relied on manually constructed executable environments. AgentGym~\citep{agentgym} unified diverse domains, $\tau$-bench~\citep{taubench} modeled stateful tool--agent--user interactions, and AppWorld~\citep{appworld} exposed controllable applications with executable evaluation. Their domain-specific interfaces, tasks, initialization, and verifiers make construction and large-scale instantiation costly. To reduce authoring cost, ScaleEnv~\citep{scaleenv} synthesized interactive environments, Gym-Anything~\citep{gymanything} converted arbitrary software into agent environments, and Agent-World~\citep{agentworld} discovered executable tool ecosystems and generated verifiable tasks. GUI-GENESIS~\citep{guigenesis}, ClawGym~\citep{clawgym}, and CUA-Gym~\citep{cuagym} similarly scaled construction for software-interacting agents. Another line reduced rollout cost through simulation. DreamGym~\citep{dreamgym} distilled environment dynamics into an experience model, while Agent World Model~\citep{agentworldmodel}, WebWorld~\citep{webworld}, Qwen-AgentWorld~\citep{qwenagentworld}, and SearchEyes~\citep{searcheyes} built learned or structured simulated worlds. Dockerless~\citep{dockerless} replaced execution-based rewards with environment-free verification. Yet simulation and execution-free feedback omit application dynamics, hidden state, and real side effects, creating a persistent sim-to-real gap; high-fidelity agent learning therefore remains dependent on costly, stateful software execution at scale.

\textbf{Agent Execution Infrastructure.} Docker~\citep{merkel2014docker} established containers as the standard substrate for reproducible, isolated execution. SOCK~\citep{sock} and SAND~\citep{sand} reduced provisioning overhead through specialization and sandbox sharing, while Firecracker~\citep{firecracker} introduced lightweight microVMs for dense isolation. For serverless functions, SEUSS~\citep{seuss} and Catalyzer~\citep{catalyzer} reused initialized state through snapshots and sandbox forking, and Groundhog~\citep{groundhog} restored a shared runtime between sequential invocations. Agent workloads, however, are long-running and stateful. WebServ~\citep{webserv} applied block-level copy-on-write to web environments, SWE-MiniSandbox~\citep{sweminisandbox} combined kernel isolation with reusable caches, and Orchard~\citep{orchard} exposed reusable sandbox-lifecycle primitives. AgentCgroup~\citep{agentcgroup}, Crab~\citep{crab}, Sandlock~\citep{wang2026sandlockconfiningaiagent}, and SpecBox~\citep{specbox} improved resource control, checkpointing, confinement, and scheduling, while the Rollout Infrastructure Tax~\citep{rolloutinfrastructuretax} quantified agent-rollout systems overhead. Yet these systems still allocate a separate runtime to each concurrent environment, whereas prior runtime sharing assumes stateless or sequential execution. Supporting concurrently diverging mutable states without replicating complete runtimes remains an open systems problem for large-scale agent training.

\section{CUA-Sandbox}
\label{sec:method}

\subsection{Overview}
\label{sec:method-overview}

CUA-Sandbox separates the mutable state of a logical environment from the runtime that executes its actions. Each environment maintains a private \emph{state capsule}, while initialized application processes are reused across environments. During execution, environment-specific bindings direct actions to the corresponding capsule. Resetting or branching an environment therefore operates on its state, while the shared runtime remains initialized.

Consider $N$ environments running the same application stack. A conventional deployment couples each environment's state to its own initialized processes. CUA-Sandbox instead makes the state and its bindings the environment-specific components:
\begin{equation}
    \mathcal{E}_i^{\mathrm{rep}}=\langle P_i,B,S_i\rangle,
    \qquad
    \mathcal{E}_i^{\mathrm{CUA}}=\langle P,B,K_i,\Phi_i\rangle.
    \label{eq:environment-representation}
\end{equation}
Here, $P_i$ denotes a private initialized runtime and $P$ the shared runtime. Immutable software and base data $B$ may already be shared in either design. Capsule $K_i$ represents trajectory state $S_i$, and $\Phi_i$ binds application resources to that capsule. The change concerns environment realization; task setup, interfaces, and evaluation remain the same.

Figure~\ref{fig:cua-overview} outlines the execution flow. The system constructs a capsule from the task's state requirements, binds an action and its resulting observations to that capsule, and coordinates state transitions when the environment is reset or branched. We describe capsule construction in Section~\ref{sec:state-capsules}, state-scoped execution in Section~\ref{sec:state-execution}, and lifecycle management in Section~\ref{sec:state-lifecycle}. These mechanisms target two requirements: mutations remain private to their environment, and execution preserves the task-relevant behavior of the underlying software.

\subsection{Environment State Capsules}
\label{sec:state-capsules}

The first step is to identify the mutable resources that determine future observations and rewards. A capsule collects these resources into one independently managed environment state, allowing the application runtime to be reused without merging trajectories.

\textbf{Identifying mutable state.} A task resource contract specifies the databases, user files, browser profiles, service state, and interface resources required by a task family. It is constructed from task setup, evaluator dependencies, and application-state specifications. For each resource, the contract selects a backend adapter and a lifecycle policy. For example, two shopping rollouts can use the same application server while resolving database and session access to separate environment state. Resources without a fine-grained adapter fall back to a coarse private backend; filesystem copy-on-write covers writes outside the fine-grained file entries. This fallback trades sharing efficiency for a larger private state boundary.

\textbf{Constructing state capsules.} We distinguish three classes of state according to how they must be preserved. \emph{Authoritative state}, such as databases and user-created files, is stored as a private delta over immutable base data. \emph{Derived state}, such as caches and search indexes, uses private namespaces and may be reconstructed when rebuilding is cheaper than copying. \emph{Ephemeral state}, such as temporary IPC endpoints and display resources, is recreated when a capsule is activated. Their backends differ, but all three belong to the same logical environment.

Each capsule has an environment identity, a generation number, and a manifest recording its component versions and dependencies. Backend adapters expose a common interface for preparation, quiescence, snapshotting, forking, resetting, activation, and destruction. The manifest commits only after all required components are ready, so a database branch, file overlay, and application profile advance as one generation. Appendix~\ref{sec:capsule-details} gives the resource and binding representations.

\subsection{State-Scoped Execution}
\label{sec:state-execution}

State capsules define what remains private; state-scoped execution determines how shared processes access it. Every action executes with an environment identity and a binding table that selects the corresponding data branches, service namespaces, and interface endpoints.

\textbf{Binding actions to state.} Before an action executes, the control plane resolves the environment's current bindings and acquires a generation lease. For authoritative resource $k$, the resulting view is
\begin{equation}
    V_{i,k}^{g}=b_k\oplus\delta_{i,k}^{g},
    \label{eq:overlay-view}
\end{equation}
where $b_k$ is immutable base data, $\delta_{i,k}^{g}$ is the private branch for environment $i$ at generation $g$, and $\oplus$ denotes backend-specific overlay composition. Reads use the composed view, while writes terminate in the private branch. Derived services and ephemeral interfaces are selected through their environment-specific namespaces and endpoints. The action's causal requests, callbacks, and background work inherit the same context; observations and evaluator access use the same generation.

\textbf{Maintaining isolation.} Runtime sharing requires complete resource bindings, disjoint writable state, and a consistent generation throughout an operation. If $W_i^g$ denotes the writable state reachable under an environment's bindings, the isolation requirement is
\begin{equation}
    W_i^g\cap W_j^{g'}=\emptyset,\qquad i\neq j.
    \label{eq:disjoint-writes}
\end{equation}
Immutable bases may be shared, but each environment's writes must resolve to its own branches and namespaces. A context becomes ready only after every contracted resource has a binding. Generation-scoped leases keep ongoing work attached to one version and reject stale requests after a transition. These requirements depend on covering the task's mutable resources and propagating context through their adapters; sharing a process alone does not establish isolation.

The binding granularity follows the application. Web services can select private state per request, while browser and desktop interactions require session-scoped profiles, configuration, display, and message-bus endpoints. Both cases expose the original interaction interface to the agent. Because execution can be asynchronous or nondeterministic, fidelity is evaluated through task-relevant observations, evaluator-visible state, and final outcomes.

\subsection{Lifecycle and Runtime Reuse}
\label{sec:state-lifecycle}

Once execution is bound to private state, lifecycle operations can replace or branch that state without rebuilding the shared runtime. The control plane coordinates these operations with ongoing execution so that actions never observe a partially updated capsule.

\textbf{State-level operations.} Reset constructs a clean successor capsule from the immutable base. Snapshot records an immutable version of the current capsule, while clone and fork create copy-on-write descendants that share unchanged data and diverge as new actions modify them. These operations preserve the initialized application processes and act on the capsule's component backends.

\textbf{Atomic transitions.} A transition freezes the environment's route and drains in-flight requests and background work. Backends stage a successor capsule and its bindings. After all components commit and pass readiness checks, the control plane atomically publishes the new generation:
\begin{equation}
    \operatorname{active}[i]\leftarrow\langle g+1,\Phi_i^{g+1}\rangle.
    \label{eq:binding-publication}
\end{equation}

\begin{table}[t]
\centering
\caption{\lead{Inference performance with matched execution backends.} Success rates (\%) are unweighted site/category means. Shaded rows show CUA-Sandbox; bold marks pairwise maxima. Site-level scores appear in Appendix~\ref{sec:detailed-results}.}
\label{tab:main-results}
\tablefont
\begin{tabularx}{\linewidth}{@{}ll*{3}{>{\centering\arraybackslash}X}@{}}
\toprule
\thead{Model} & \thead{Backend} & \thead{WebArena} & \thead{VisualWebArena} & \thead{OSWorld} \\
\midrule
GPT-5.6-Terra & Docker & 21.89 & 28.34 & 67.29 \\
\rowcolor{ResultOurs}  & CUA-Sandbox & \textbf{22.43} & \textbf{29.06} & \textbf{70.62} \\
\midrule
Qwen3.5-9B & Docker & 9.67 & 10.52 & 24.17 \\
\rowcolor{ResultOurs}  & CUA-Sandbox & \textbf{10.49} & \textbf{13.54} & \textbf{27.66} \\
\midrule
Muse-Glimmer-30B & Docker & 23.14 & 25.36 & 24.40 \\
\rowcolor{ResultOurs}  & CUA-Sandbox & \textbf{25.63} & \textbf{25.80} & \textbf{34.39} \\
\midrule
Gemma-4-31B-IT & Docker & 24.97 & 19.07 & 32.61 \\
\rowcolor{ResultOurs}  & CUA-Sandbox & \textbf{26.34} & \textbf{20.36} & \textbf{35.54} \\
\bottomrule
\end{tabularx}
\end{table}
\begin{table}[t]
\centering
\caption{\lead{Qwen3.5-4B performance before and after training.} Success rates (\%) follow Base $\rightarrow$ SFT $\rightarrow$ RL under each training backend. Final checkpoints share the reference evaluation pipeline. Bold marks final RL scores; $\Delta$ is the RL gain over SFT in percentage points.}
\label{tab:training-results}
\tablefont
\begin{tabularx}{\linewidth}{@{}ll*{4}{>{\centering\arraybackslash}X}@{}}
\toprule
\thead{Benchmark} & \thead{Training backend} & \thead{Base} & \thead{SFT} & \thead{RL} & \thead{$\Delta$ (pp)} \\
\midrule
WebArena & Docker & 8.25 & 9.85 & \textbf{12.93} & +3.08 \\
\rowcolor{ResultOurs}  & CUA-Sandbox & 8.25 & 10.10 & \textbf{13.05} & +2.95 \\
\midrule
VisualWebArena & Docker & 18.13 & 23.84 & \textbf{27.03} & +3.19 \\
\rowcolor{ResultOurs}  & CUA-Sandbox & 18.13 & 24.07 & \textbf{27.03} & +2.96 \\
\midrule
OSWorld & Docker & 14.96 & 21.61 & \textbf{25.48} & +3.87 \\
\rowcolor{ResultOurs}  & CUA-Sandbox & 14.96 & 22.16 & \textbf{26.04} & +3.88 \\
\bottomrule
\end{tabularx}
\end{table}
A failed transition retains the previous published generation. Together with generation-scoped leases, this protocol prevents delayed actions or evaluator requests from crossing a state transition and prevents execution against mixed component versions.

\textbf{Reusing execution slots.} The number of materialized capsules need not equal the number of active execution contexts. CUA-Sandbox maintains $N$ logical environments and allocates $Q$ physical execution slots as needed; inactive environments remain as capsules. Rebinding a slot quiesces its current context, attaches the target capsule, recreates ephemeral resources, and updates the route before passing a readiness barrier. A slot therefore has no permanent environment identity. The state plane manages capsules, the execution plane applies their bindings, and the control plane schedules slots and publishes lifecycle transitions. A memory and storage model appears in Appendix~\ref{sec:scaling-analysis}.

\begin{figure}[t]
\centering
\includegraphics[width=\linewidth]{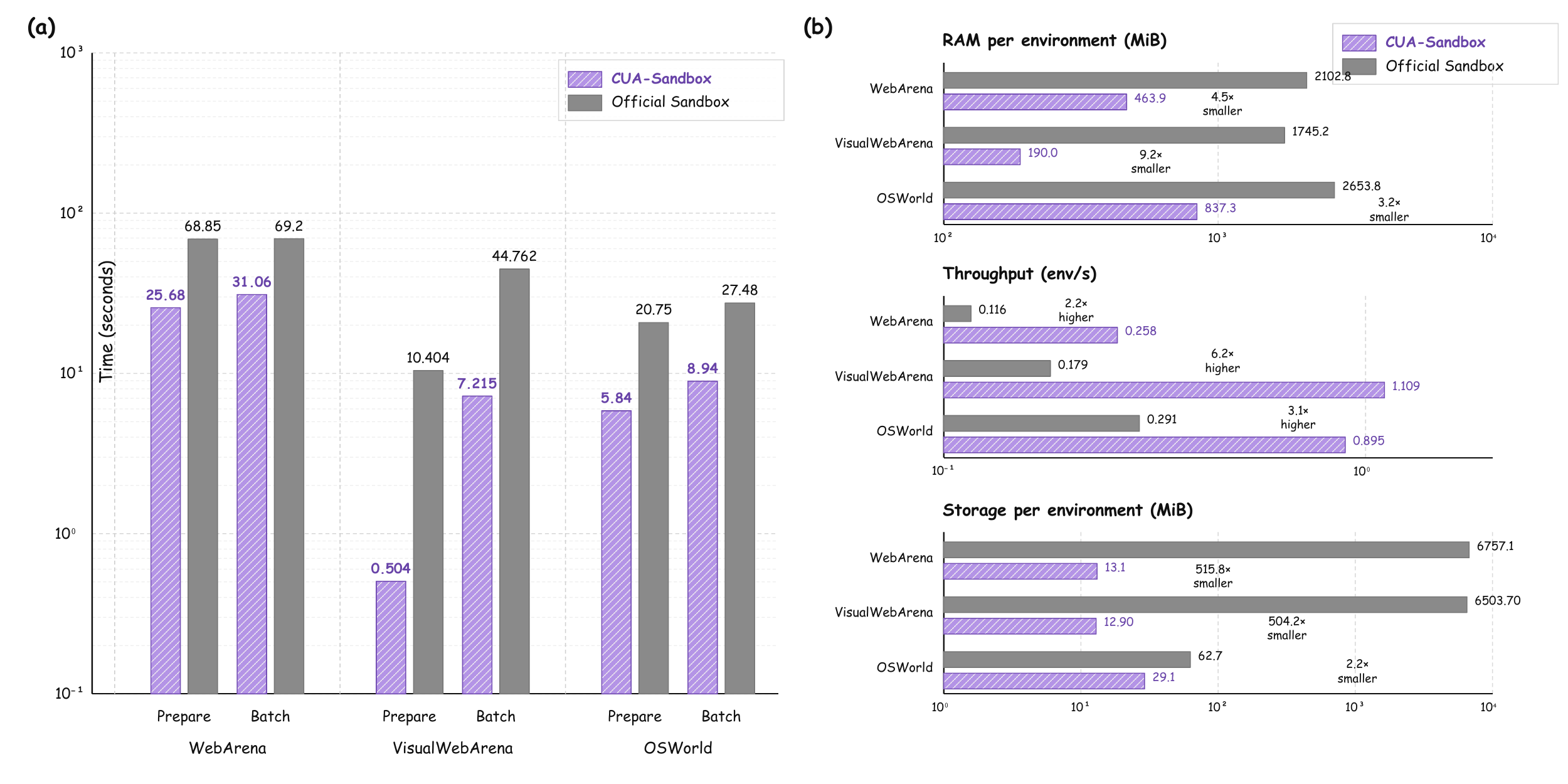}
\caption{\lead{Rollout efficiency with eight concurrent environments.} (a) Preparation and batch wall-clock time. (b) Per-environment memory, throughput, and incremental storage. Purple hatched bars show CUA-Sandbox; gray bars show the Docker baseline (Official Sandbox). Axes are logarithmic. Lower time and resource costs and higher throughput are better. Table~\ref{tab:rollout-absolute} lists the measurements.}
\label{fig:rl-efficiency}
\end{figure}

\interlinepenalty=10000
\section{Experiments}
We evaluate whether runtime sharing preserves agent performance, improves rollout efficiency, and reduces environment costs. We then ablate the state and lifecycle mechanisms to separate their contributions. Tables~\ref{tab:main-results}--\ref{tab:system_efficiency} and Figures~\ref{fig:rl-efficiency}--\ref{fig:ablations} address these questions; detailed evaluation settings and scaling analysis are deferred to the appendix.

\subsection{Experimental Setup}
\label{sec:main-experimental-setup}

\textbf{Benchmarks and models.} We evaluate WebArena, VisualWebArena, and OSWorld using their original task interfaces and evaluators~\citep{zhou2024webarena,koh2024visualwebarena,xie2024osworld}. The inference study uses GPT-5.6-Terra, Qwen3.5-9B, Muse-Glimmer-30B, and Gemma-4-31B-IT. Post-training uses WebArena, VisualWebArena, and OSWorld, with Qwen3.8-27B as the teacher and Qwen3.5-4B as the student. 

\textbf{Backends and controls.}
We compare CUA-Sandbox with the conventional Docker-based environment for
each benchmark under matched experimental conditions. Each inference pair
uses the same model, prompts, decoding settings, interaction horizon, task
interface, and official evaluator, so that the execution backend is the
primary experimental variable. For post-training, both comparisons start
from the same Qwen3.5-4B base checkpoint. Successful teacher trajectories
are collected independently under each backend to construct the corresponding
SFT data, after which on-policy RL is performed using the same backend as
trajectory collection. Across the two training runs, we match the data and
rollout budgets, task sampling procedure, decoding configuration,
optimization hyperparameters, reward function, and interaction horizon.
Final checkpoints are evaluated through the same reference evaluation
pipeline, ensuring that differences in measured task performance are not
introduced by changes in prompts, evaluators, or training budgets.

\textbf{Metrics and compute.} We report task success and its unweighted mean across sites or task categories, keeping each benchmark separate. The rollout study fixes concurrency at eight environments and measures preparation time, batch wall-clock time, throughput, memory, and incremental storage. Operation-level measurements separately cover resource use and five lifecycle operations. All runs use a host with 8 NVIDIA RTX PRO 5000 Blackwell GPUs (72~GB each), 192 physical CPU cores, 1.5~TiB memory, and two approximately 3.5~TB NVMe drives, running CUDA~13.0. Appendix~\ref{sec:experimental-setup} gives decoding and evaluation details.

\subsection{Agent Performance}

\textbf{Runtime sharing preserves task performance across agents.} CUA-Sandbox scores above Docker in all twelve model--benchmark pairs (Table~\ref{tab:main-results}). For GPT-5.6-Terra, success increases from 21.89\% to 22.43\% on WebArena-Lite, from 28.34\% to 29.06\% on VisualWebArena, and from 67.29\% to 70.62\% on OSWorld. The same direction holds for Qwen3.5-9B, Muse-Glimmer-30B, and Gemma-4-31B-IT despite their different starting scores. This consistency supports runtime reuse across agent families and interaction interfaces under the tested prompts and evaluators.

\textbf{The efficiency gains carry through agent post-training.} The CUA-Sandbox student finishes 0.12 percentage points above Docker on WebArena, matches it on VisualWebArena, and finishes 0.56 points above it on OSWorld (Table~\ref{tab:training-results}). OSWorld provides the closest comparison of the RL stage: success rises from 22.16\% to 26.04\% with CUA-Sandbox and from 21.61\% to 25.48\% with Docker, gains of 3.88 and 3.87 points. Together with the web results, these matched runs show that the cheaper execution backend retains the measured benefit of on-policy interaction.

\subsection{Rollout Efficiency}

\begin{table}[t]
\centering
\caption{\lead{Where runtime sharing saves resources and lifecycle work.} Entries are conventional-environment costs divided by CUA-Sandbox costs ($\times$); larger is better. The first two rows report resource reductions, and the remaining rows report operation speedups. Shaded bold cells mark the largest factor per metric.}
\label{tab:system_efficiency}
\tablefont
\setlength{\tabcolsep}{3pt}
\begin{tabularx}{\linewidth}{@{}l*{7}{>{\centering\arraybackslash}X}@{}}
\toprule
 & \multicolumn{3}{c}{\thead{WebArena}} & \multicolumn{3}{c}{\thead{VisualWebArena}} & \thead{OSWorld} \\
\cmidrule(lr){2-4}\cmidrule(lr){5-7}\cmidrule(lr){8-8}
\thead{Metric} & \shortstack{Shopping\\Admin} & Shopping & GitLab & Shopping & Reddit & Classifieds & Desktop \\
\midrule
Memory & 20.6 & 46.5 & \cellcolor{ResultOurs}\textbf{51.2} & 1.92 & 22.0 & 3.70 & 1.45 \\
Storage & 60.1 & 543.6 & 6.3 & 391.2 & \cellcolor{ResultOurs}\textbf{9,535} & 3.09 & 3.67 \\
\midrule
Create & 1.70 & 1.15 & 8.82 & \cellcolor{ResultOurs}\textbf{15.17} & 8.75 & 6.89 & 2.98 \\
Reset & 1.20 & 1.03 & 5.51 & 4.99 & \cellcolor{ResultOurs}\textbf{7.29} & 2.66 & 3.17 \\
Restore & 1.20 & 1.21 & 6.62 & 5.78 & \cellcolor{ResultOurs}\textbf{7.39} & 2.52 & 2.90 \\
Fork & 1.10 & 1.25 & 9.37 & \cellcolor{ResultOurs}\textbf{17.84} & 10.18 & 6.65 & 2.98 \\
Clone & 2.13 & 1.17 & \cellcolor{ResultOurs}\textbf{7.47} & 2.10 & 1.96 & 3.30 & 1.93 \\
\bottomrule
\end{tabularx}
\end{table}

\begin{figure}[t]
\centering
\input{ablation-figure.tex}
\caption{\lead{Component ablations separate efficiency costs from lifecycle correctness.} Thin purple bars show CUA-Sandbox; wide gray bars show the named ablation. Bar lengths use a shared linear scale within each panel; labels give exact values. (a) Copy-on-write on a 3.6~GB Shopping capsule. (b) Non-DB state management on read-only WebArena task~21. (c) Publication events over 100 trials; purple origin markers denote zero events, and N/A denotes an inapplicable comparison.}
\label{fig:ablations}
\end{figure}
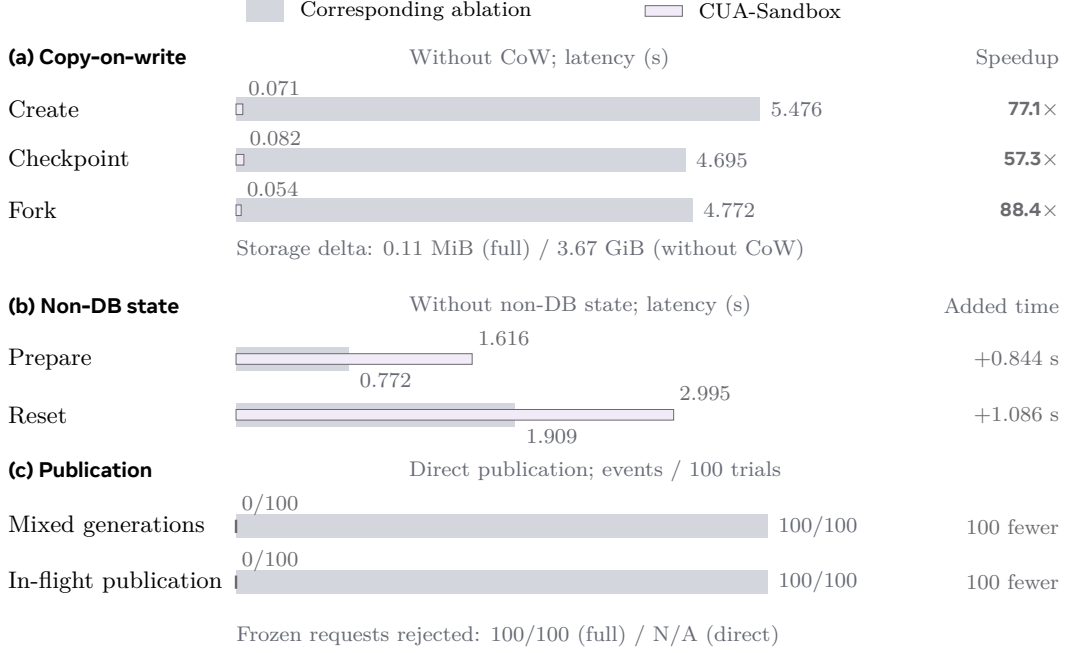

\textbf{Preparation savings increase end-to-end rollout throughput.}
VisualWebArena shows the largest throughput gain: preparation falls from
10.404 to 0.504~s, batch time falls from 44.762 to 7.215~s, and throughput
rises from 0.179 to 1.109 environments/s
(Figure~\ref{fig:rl-efficiency}). The 20.64$\times$ preparation speedup
becomes a 6.20$\times$ batch speedup, showing that initialization accounts
for only part of the rollout cost. Throughput also increases by 2.22$\times$
on WebArena and 3.08$\times$ on OSWorld. Runtime reuse therefore improves
the complete measured batch, beyond the environment setup step.

\textbf{Higher throughput comes with a smaller per-environment footprint.}
On WebArena, memory falls from 2,102.8 to 463.9~MiB while incremental storage
falls from 6,757.1 to 13.1~MiB. VisualWebArena shows a similarly strong
reduction, with memory falling from 1,745.2 to 190.0~MiB and incremental
storage falling from 6,503.70 to 12.90~MiB. OSWorld memory falls from
2,653.8 to 837.3~MiB, while incremental storage falls from 62.7 to
29.1~MiB. The larger web storage savings are consistent with sharing
initialized application state that would otherwise be materialized per
environment. These results concern eight concurrent environments; maximum
density remains unmeasured. 

\subsection{Environment Resource and Lifecycle Costs}

\textbf{Memory and storage savings peak on different workloads.} WebArena GitLab reduces memory by 51.2$\times$ but storage by 6.3$\times$; VisualWebArena Reddit reduces memory by 22.0$\times$ and storage by 9,535$\times$ (Table~\ref{tab:system_efficiency}). A workload with the largest memory benefit need not have the largest storage benefit. The two metrics capture different costs of maintaining independent environments, and the private state that remains after runtime sharing depends on the application.

\textbf{Lifecycle speedups depend on both the operation and the application.} VisualWebArena Shopping achieves a 15.17$\times$ create speedup and a 17.84$\times$ fork speedup, whereas WebArena Shopping improves by 1.15$\times$ and 1.25$\times$ on the same operations. Its reset speedup is only 1.03$\times$. These differences delimit the benefit of runtime reuse: avoiding process initialization helps most where it dominates the original operation, while capsule preparation and readiness checks still incur latency.

\subsection{Ablation Study}
\label{sec:ablation}
We remove copy-on-write, non-DB state management, and transactional publication in turn to separate their effects on cost and correctness. Appendix~\ref{sec:ablation-details} provides controls and measurement details, including task selection, readiness checks, and stress-test conditions. More experients are provided in Appendix~\ref{sec:ablation33}

\textbf{Copy-on-write removes the dominant state-materialization cost.} With the same 3.6~GB Shopping capsule, removing copy-on-write raises create latency from 0.071 to 5.476~s and fork latency from 0.054 to 4.772~s (Figure~\ref{fig:ablations}). The storage delta rises from 0.11~MiB to 3.67~GiB. Because the capsule and lifecycle operations are otherwise unchanged, this ablation identifies state materialization as the principal cost avoided by copy-on-write.

\textbf{Managing non-DB state adds adapter overhead.} On WebArena task~21, including non-DB state adds 0.844~s to preparation and 1.086~s to reset; all four HTTP readiness probes return status~200. This cost is the work required to manage resources beyond the database. The task is read-only, so the comparison isolates adapter overhead rather than isolation under state-changing actions.

\textbf{Transactional publication prevents mixed-generation execution.} In 100 stress-test trials, the full protocol exposes no mixed generations and never publishes while requests remain in flight. Direct publication produces both events in all 100 trials. The full protocol also rejects all 100 requests issued while the route is frozen. These controls show why state reuse requires a coordinated transition: successor state must become visible only after ongoing execution has drained.
\FloatBarrier
\interlinepenalty=0
\section{Conclusion}
CUA-Sandbox separates the mutable state of computer-use environments from their initialized execution runtime. State capsules, environment-scoped bindings, and transactional lifecycle operations enable reuse while retaining the original software interfaces and evaluators. Across web and desktop workloads, the measured resource and lifecycle savings translate into higher rollout throughput, with maintained task performance and similar post-training improvements. The results support runtime reuse as a practical systems component for agent training. The ablations clarify this tradeoff: copy-on-write avoids state materialization, while non-database adapters incur measurable overhead and transactional publication protects lifecycle consistency. The approach therefore depends on complete resource contracts and binding coverage, with private backends retained wherever safe sharing cannot be established. Broader mutation workloads, longer trajectories, and concurrency sweeps are needed to characterize the remaining sharing boundary, stress state isolation under diverse write patterns, and determine how these gains translate into achievable environment density.

\clearpage
\subsection*{AI Use Statement}
AI tools assisted with language editing, manuscript organization, and LaTeX presentation.

\bibliography{paper}
\bibliographystyle{assets/plainnat}
\clearpage
\appendix
\section{Supplementary Ablations and Execution Configurations}
\label{sec:supplementary-ablations}
We first report the controls and interpretation of the component ablations in Figure~\ref{fig:ablations}, then compare additional execution configurations. Each ablation removes one mechanism from complete CUA-Sandbox; the separate Docker comparison evaluates the execution backend as a whole.

\subsection{Component Ablation Details}
\label{sec:ablation-details}
\textbf{Copy-on-write.} Both configurations use the same 3.6~GB Shopping MySQL capsule. We measure create, checkpoint, and fork latency, together with the physical storage delta for a capsule operation. Removing copy-on-write raises the three latencies from 0.071, 0.082, and 0.054~s to 5.476, 4.695, and 4.772~s, respectively. Storage grows from 0.11~MiB to 3.67~GiB. Figure~\ref{fig:ablations} computes speedups from these reported latency values; storage retains its original units.

\textbf{Non-database state.} The comparison uses WebArena task~21 and removes non-DB state management while retaining the rest of the system. Preparation takes 1.616~s with the complete system and 0.772~s after removal; reset takes 2.995 and 1.909~s. The resulting overheads are 0.844 and 1.086~s, and all four HTTP readiness probes return status~200. Because this task is read-only, the comparison measures adapter overhead rather than isolation during state-changing interaction.

\textbf{Transactional publication.} Over 100 stress-test trials, we compare the full freeze/drain and atomic-publication protocol with direct publication. The complete protocol yields no mixed-generation events and no publication while requests remain in flight; direct publication yields both events in every trial. The full protocol also rejects all 100 requests issued while the route is frozen. Frozen-request rejection does not apply to direct publication and is reported as N/A.

\subsection{Additional Execution Configurations}
\label{sec:additional-configurations}
Table~\ref{tab:absolute-costs} complements the component ablations with absolute launch, storage, and memory measurements for WebServ Incus, fresh Docker, and DB/reflink configurations. Launch latency is averaged over eight subsequent runs. DB/reflink reports marginal memory growth, while the other entries report working-set memory; the two quantities are not directly comparable.
\begin{table}[H]
\centering
\caption{\lead{Absolute costs of additional execution configurations.} Launch latency is the mean of eight subsequent runs. Storage is per environment; $^{\dagger}$ marks marginal memory growth, while the other memory entries are working-set measurements.}
\label{tab:absolute-costs}
\tablefont
\small\setlength{\tabcolsep}{4pt}
\begin{tabularx}{\linewidth}{@{}l*{3}{>{\centering\arraybackslash}X}@{}}
\toprule
\thead{Configuration} & \thead{Launch (s)} & \thead{Storage} & \thead{Memory (GiB)} \\
\midrule
WebServ Incus & 1.781 & 28.01 MiB & 1.74 \\
\rowcolor{ResultOurs} CUA-Sandbox: fresh Docker & 12.323 & 6.621 GiB & 1.795 \\
\rowcolor{ResultOurs} CUA-Sandbox: DB/reflink & 1.730 & 12.50 MiB & 0.201$^{\dagger}$ \\
\bottomrule
\end{tabularx}

\end{table}
\FloatBarrier
\clearpage
\section{Detailed Experimental Results}
\label{sec:detailed-results}
\subsection{Task Success by Site and Category}
Table~\ref{tab:detailed-results} reports the paired site and task-category scores used to compute the benchmark means in Table~\ref{tab:main-results}. Each entry retains the same model and evaluation protocol across the two backends.
\begin{table}[H]
\centering
\caption{\lead{Detailed task success by site and category.} Each cell reports Docker / \textbf{CUA-Sandbox} success rates (\%) for the same model and evaluation protocol. Benchmark means appear in Table~\ref{tab:main-results}; bold distinguishes the CUA-Sandbox result in each pair.}
\label{tab:detailed-results}
\tablefont
\setlength{\tabcolsep}{3pt}
\begin{tabularx}{\linewidth}{@{}l*{3}{>{\centering\arraybackslash}X}@{}}
\toprule
\multicolumn{4}{l}{\thead{(a) WebArena-Lite}} \\
\thead{Model} & Shopping & CMS & GitLab \\
\midrule
GPT-5.6-Terra & 20.86\,/\,\textbf{21.98} & 20.90\,/\,\textbf{20.90} & 23.90\,/\,\textbf{24.40} \\
Qwen3.5-9B & 8.02\,/\,\textbf{8.79} & 9.89\,/\,\textbf{9.89} & 11.11\,/\,\textbf{12.78} \\
Muse-Glimmer-30B & 24.60\,/\,\textbf{27.22} & 17.03\,/\,\textbf{18.13} & 27.78\,/\,\textbf{31.55} \\
Gemma-4-31B-IT & 25.13\,/\,\textbf{27.81} & 20.88\,/\,\textbf{22.32} & 28.89\,/\,\textbf{28.89} \\
\bottomrule
\end{tabularx}
\par\vspace{7pt}
\begin{tabularx}{\linewidth}{@{}l*{3}{>{\centering\arraybackslash}X}@{}}
\toprule
\multicolumn{4}{l}{\thead{(b) VisualWebArena}} \\
\thead{Model} & Classifieds & Reddit & Shopping \\
\midrule
GPT-5.6-Terra & 27.35\,/\,\textbf{27.78} & 25.71\,/\,\textbf{25.71} & 31.97\,/\,\textbf{33.69} \\
Qwen3.5-9B & 13.18\,/\,\textbf{15.38} & 5.71\,/\,\textbf{5.71} & 12.66\,/\,\textbf{19.53} \\
Muse-Glimmer-30B & 23.93\,/\,\textbf{24.78} & 19.52\,/\,\textbf{20.00} & 32.62\,/\,\textbf{32.62} \\
Gemma-4-31B-IT & 19.66\,/\,\textbf{21.37} & 15.24\,/\,\textbf{16.32} & 22.32\,/\,\textbf{23.39} \\
\bottomrule
\end{tabularx}
\par\vspace{7pt}
\begin{tabularx}{\linewidth}{@{}l*{5}{>{\centering\arraybackslash}X}@{}}
\toprule
\multicolumn{6}{l}{\thead{(c) OSWorld}} \\
\thead{Model} & OS & Office & Daily & Professional & Workflow \\
\midrule
GPT-5.6-Terra & 75.00\,/\,\textbf{83.33} & 72.36\,/\,\textbf{74.30} & 56.16\,/\,\textbf{57.42} & 79.59\,/\,\textbf{81.64} & 53.34\,/\,\textbf{56.43} \\
Qwen3.5-9B & 25.00\,/\,\textbf{29.17} & 20.43\,/\,\textbf{23.11} & 37.16\,/\,\textbf{37.56} & 30.61\,/\,\textbf{36.73} & 7.65\,/\,\textbf{11.72} \\
Muse-Glimmer-30B & 45.83\,/\,\textbf{50.00} & 12.58\,/\,\textbf{16.15} & 28.21\,/\,\textbf{34.81} & 24.49\,/\,\textbf{53.06} & 10.87\,/\,\textbf{17.94} \\
Gemma-4-31B-IT & 70.83\,/\,\textbf{70.83} & 7.76\,/\,\textbf{8.55} & 45.35\,/\,\textbf{45.35} & 22.45\,/\,\textbf{34.69} & 16.68\,/\,\textbf{18.28} \\
\bottomrule
\end{tabularx}
\end{table}
\subsection{Absolute Rollout Measurements}
Table~\ref{tab:rollout-absolute} provides the measurements underlying Figure~\ref{fig:rl-efficiency} at eight-way concurrency. Resource costs are per environment; throughput and batch time describe the complete concurrent workload.
\begin{table}[H]
\centering
\caption{\lead{Exact rollout measurements for Figure~\ref{fig:rl-efficiency}.} All settings use eight concurrent environments. Memory and incremental storage are per environment; CUA denotes CUA-Sandbox. Preparation and batch wall-clock time are in seconds.}
\label{tab:rollout-absolute}
\tablefont
\setlength{\tabcolsep}{3.5pt}
\begin{tabular}{@{}>{\raggedright\arraybackslash}p{99pt}*{3}{>{\centering\arraybackslash}p{41pt}>{\columncolor{ResultOurs}\centering\arraybackslash}p{41pt}}@{}}
\toprule
\thead{Metric}  & \multicolumn{2}{c}{\thead{WebArena}} & \multicolumn{2}{c}{\thead{VisualWebArena}} & \multicolumn{2}{c}{\thead{OSWorld}} \\
\cmidrule(lr){2-3}\cmidrule(lr){4-5}\cmidrule(lr){6-7}
 & Docker & \cellcolor{ResultOurs}\thead{CUA} & Docker & \cellcolor{ResultOurs}\thead{CUA} & Docker & \cellcolor{ResultOurs}\thead{CUA} \\
\midrule
Prepare (s) & 68.85 & 25.68 & 6.964 & 0.633 & 20.75 & 5.84 \\
Batch (s) & 69.20 & 31.06 & 9.104 & 2.138 & 27.48 & 8.94 \\
Throughput (env/s) & 0.116 & 0.258 & 0.220 & 0.936 & 0.291 & 0.895 \\
Memory (MiB) & 2102.8 & 463.9 & 328.9 & 188.4 & 2653.8 & 837.3 \\
Storage (MiB) & 6757.1 & 13.1 & 4879.05 & 12.52 & 62.7 & 29.1 \\
\bottomrule
\end{tabular}
\end{table}
\FloatBarrier
\section{Experimental Setup and Evaluation Protocols}

\subsection{Model Settings}
\label{sec:experimental-setup}

During inference, Qwen3.5-9B used \texttt{enable\_thinking=true},
temperature $1.0$, top-$p=0.9$, and 512 maximum output tokens for OSWorld,
and temperature $1.0$, top-$p=0.9$, and 384 maximum output tokens for web
tasks.  Gemma-4-31B-IT used \texttt{enable\_thinking=true}, temperature
$1.0$, top-$p=0.95$, and 4096 maximum output tokens for OSWorld or 1000 for
web tasks.  GPT-5.6-Terra used temperature $1.0$, top-$p=0.9$, 1000 maximum
output tokens, and \texttt{reasoning\_effort=low}.  Muse-Glimmer-30B used
temperature $1.0$, top-$p=0.9$, and an output limit of 1000 tokens.

\subsection{More Details}
\label{sec:experimental-details}

To evaluate the framework itself, we first focus on the supervised fine-tuning (SFT) stage. We use Qwen3.8-27B as the teacher model to generate 5k successful trajectories from the same training-task pool under both the conventional sandbox and CUA-Sandbox. We retain trajectories that successfully pass the corresponding task evaluator and use them to independently fine-tune Qwen3.5-4B for 2 epochs, obtaining one SFT checkpoint for each execution environment.
Next, we perform on-policy RL training starting from the corresponding SFT checkpoints using 1,600 training instances for 1 epoch under both environment backends. We set the RL batch size to 16 and the rollout count to 8, resulting in 128 parallel logical environments per update. Both runs use the same task sampler, decoding configuration, maximum interaction horizon, reward function, and optimization hyperparameters, with the execution environment being the only difference.

\subsection{Benchmarks and Evaluation Protocols}

\paragraph{WebArena.}
WebArena~\cite{zhou2024webarena} is a realistic and reproducible benchmark for evaluating autonomous web agents on long-horizon tasks. It provides fully functional, self-hosted web applications spanning e-commerce, social discussion, collaborative software development, and content management, together with auxiliary tools and knowledge resources. The benchmark contains 812 natural-language tasks and evaluates agents based on the functional correctness of the resulting website state rather than exact action-sequence matching. In our experiments, we focus on the Shopping, CMS, and GitLab domains, which involve state-changing interactions and therefore require isolated and resettable execution environments.
\paragraph{VisualWebArena.}
VisualWebArena~\cite{koh2024visualwebarena} extends web-agent evaluation to visually grounded tasks that require joint reasoning over images and textual webpage content. It contains 910 tasks across three self-hosted environments: Classifieds, Shopping, and Reddit. The Shopping and Reddit environments are inherited from WebArena, while Classifieds introduces an additional visually rich domain. Completing these tasks requires agents to interpret natural-language instructions, perceive visual information on webpages, and execute multi-step browser actions. We evaluate CUA-Sandbox on all three domains using the standard multimodal interaction protocol.
\paragraph{OSWorld.}
OSWorld~\cite{xie2024osworld} is a benchmark for evaluating multimodal agents on open-ended computer-use tasks in real desktop environments. It provides a scalable execution environment with task initialization and execution-based evaluation, and includes 369 tasks involving real web and desktop applications, operating-system file operations, and cross-application workflows. Each task specifies an initial environment state together with a task-specific evaluator that directly checks the resulting system state. In our experiments, we report results across the OS, Office, Daily, Professional, and Workflow categories.
\paragraph{Agents and observations.}
WebArena agents receive the
task instruction and the accessibility tree.  VisualWebArena agents receive
the current page screenshot, the task-image captions, and the screenshot with
Set-of-Marks (SoM) element identifiers (i.e., the Image + Captions + SoM
configuration), and emit SoM-based actions.  OSWorld agents receive the task
instruction and the current desktop screenshot and interact through the
benchmark's mouse, keyboard, and terminal action interfaces.

\paragraph{Execution and compute.}
All experiments are run on the host described in Section~\ref{sec:main-experimental-setup}: 8 NVIDIA RTX PRO 5000 Blackwell GPUs (72~GB each), 192 physical CPU cores, and 1.5~TiB of system memory.
We use the benchmark defaults for
the maximum interaction horizon (30 steps for WebArena and VisualWebArena)
and keep sampling parameters, context limits, image limits, and timeout
values fixed within each benchmark.  Evaluation workers may run in parallel,
but the concurrency limit is held constant across the two environment
implementations so that the comparison does not conflate isolation with
scheduling capacity.

\paragraph{Metrics and failure handling.}
We first compute task success within each benchmark site or task category.
For group $g$, containing the evaluated task set $\mathcal{T}_g$, the
group-level success rate is
\[
\mathrm{SR}_g
=
\frac{
\#\,\{\text{successful evaluated tasks in } g\}
}{
\#\,\{\text{evaluated tasks in } g\}
}.
\]
The benchmark-level score reported in the main tables is the unweighted
mean over the corresponding sites or task categories:
\[
\mathrm{BenchmarkScore}
=
\frac{1}{|\mathcal{G}|}
\sum_{g\in\mathcal{G}} \mathrm{SR}_g,
\]
where $\mathcal{G}$ is the set of reported sites or categories.
Thus, each site or category contributes equally to the benchmark-level
score, irrespective of the number of tasks it contains. This is a macro
average over groups rather than a task-weighted micro average.

For WebArena, $\mathcal{G}$ consists of Shopping, CMS, and GitLab; for
VisualWebArena, it consists of Classifieds, Reddit, and Shopping; and for
OSWorld, it consists of OS, Office, Daily, Professional, and Workflow.
The corresponding group-level scores are reported separately in
Appendix~B.

For WebArena and VisualWebArena, the official task-specific program/HTML,
string-match, and image-based evaluators are used as applicable; for
OSWorld, we use the official task evaluator. The same judge model and
endpoint are used for both sides of a pair.

A transport retry may re-execute a crashed worker, but it does not create
an additional task or otherwise change the evaluation set. Unresolved
environment, browser, or evaluator errors for an evaluated task are
recorded as failures. Tasks excluded from evaluation are reported
separately and are not assigned a synthetic score.

In addition to task success, we record environment preparation time,
per-step environment communication time, timeout counts, and the physical
storage occupied by each environment's mutable state. Unless otherwise specified, each reported measurement is repeated five
times under the same experimental configuration, and we report the
arithmetic mean across the five runs.
\section{Implementation Details}
\label{sec:capsule-details}

\paragraph{Resource contracts.}
For a task family $\tau$, let $\mathcal{L}_\tau$ denote its logical
resource keys. The contract
\[
\Gamma_\tau
=
\{(k,c_k,a_k,p_k)\mid k\in\mathcal{L}_\tau\}
\]
associates each key with a state class $c_k$, backend adapter $a_k$, and
lifecycle policy $p_k$. Keys identify databases, file trees, profiles,
queues, displays, message-bus sessions, or other task-relevant resources.
Agents operate only on these logical keys and do not observe the associated
physical backend names.

The resource contract defines the sharing boundary conservatively. A
task-relevant mutable resource is eligible for runtime sharing only when its
accesses can be completely resolved through an environment-scoped binding.
Resources for which such a binding cannot be established are assigned a
private backend or a coarser private execution context. They are never forced
into the shared runtime merely to increase reuse. Consequently, extending
CUA-Sandbox to a setting with additional mutable resources may reduce the
amount of sharing, but does not require weakening the environment-isolation
contract.

\paragraph{Capsule metadata.}
A capsule generation is represented as
\[
K_i^g
=
\left\langle
I_i^g,
\Delta_i^{\mathrm{auth}},
N_i^{\mathrm{der}},
P_i^{\mathrm{eph}},
M_i^g
\right\rangle .
\]
Here, $I_i^g$ records the environment, branch, and generation identity;
$\Delta_i^{\mathrm{auth}}$ holds authoritative state deltas;
$N_i^{\mathrm{der}}$ records private namespaces and reconstruction rules;
$P_i^{\mathrm{eph}}$ describes ephemeral resources to recreate; and
$M_i^g$ records component versions and dependencies. The composite manifest
coordinates their publication as a single logical generation.

This representation separates persistent task state from derived and
ephemeral execution state without assuming that all mutable state is
shareable. State that can affect future observations, evaluator-visible
results, or externally visible side effects must either be represented in
the capsule, explicitly scoped by the active execution context, or retained
within a private component.

\paragraph{Binding tables.}
Each published generation has one binding for every contracted key:
\[
\operatorname{dom}(\Phi_i^g)=\mathcal{L}_\tau.
\]
An entry
\[
\Phi_i^g(k)
=
\left\langle
b_k,
\delta_{i,k}^g,
n_{i,k}^g,
e_{i,k}^g
\right\rangle
\]
selects immutable base data, a private data branch, a private service
namespace, and a private interface endpoint; fields that do not apply are
null.

The trusted execution context combines the environment identity with this
table, and a generation-scoped lease pins the selected version for the
duration of execution. A context becomes executable only after all
task-relevant contracted resources have valid bindings. Thus, for writable
state $W_i^g$ reachable by environment $i$ at generation $g$, the required
isolation invariant is
\[
W_i^g \cap W_j^{g'} = \varnothing,
\qquad i\neq j .
\]

Process sharing alone is therefore insufficient to establish isolation.
Any task-relevant mutable process state must either be keyed by the active
environment context or remain inside a private execution component. If this
condition cannot be established for a particular application component, the
component is conservatively excluded from the shared boundary.

Web adapters select databases, files, queues, and caches; visual-web
adapters additionally bind browser profiles and sessions; desktop adapters
bind filesystem, application-profile, configuration, display, message-bus,
and window state. These benchmark-specific adapters differ only in which
resources are included in the contract. The isolation rule itself is
unchanged across settings: all task-relevant writable state remains private
to one logical environment generation, while only resources satisfying the
sharing contract may be reused.

\paragraph{Asynchronous and background execution.}
Environment identity and generation apply to the complete causal execution
context rather than only to the foreground agent action. Asynchronous
requests, callbacks, and tracked background work spawned by an admitted
operation inherit the same environment context and generation lease and
therefore resolve state through the same binding table.

During reset, clone, fork, or slot rebinding, new work for the affected
environment is first quiesced and existing in-flight work is drained before
a successor generation is published. Delayed work carrying a stale
generation cannot be rebound to the successor state. Background activity
that cannot be reliably attributed, scoped, or quiesced is instead placed
outside the shared execution boundary and uses a private or coarser-grained
backend.

Thus, unsupported asynchronous behavior or application-specific state can
reduce the achievable degree of runtime reuse, but does not require
relaxing the isolation invariant.

\section{Scaling Analysis}
\label{sec:scaling-analysis}

The separation of logical environments from execution slots gives a resource model for runtime reuse. The following accounting compares the two designs under the same logical state and separates shared initialization from private execution and capsule costs.

\textbf{Memory.} Let $m_P$ denote the proportional-set memory of initialized processes and the common runtime, $m_S$ the private memory of an active slot, $m(K_i)$ the resident memory of capsule $i$, and $m_{\mathrm{ctrl}}$ the control-plane overhead. Excluding fixed host-OS memory common to both designs,
\begin{equation}
\begin{aligned}
    \mathcal{M}_{\mathrm{rep}}(N)
        &=N(m_P+m_S)+\sum_{i=1}^{N}m(K_i),\\
    \mathcal{M}_{\mathrm{CUA}}(N,Q)
        &=m_P+Qm_S+\sum_{i=1}^{N}m(K_i)+m_{\mathrm{ctrl}}.
\end{aligned}
\label{eq:memory-model}
\end{equation}
Under this decomposition, the avoided replication is
\begin{equation}
    \mathcal{M}_{\mathrm{rep}}-\mathcal{M}_{\mathrm{CUA}}
    =(N-1)m_P+(N-Q)m_S-m_{\mathrm{ctrl}}.
    \label{eq:memory-saving}
\end{equation}
The first term captures shared runtime initialization; the second captures the additional benefit when some trajectories remain inactive. Even when $Q=N$, runtime sharing can reduce memory, while each environment still incurs private state and execution costs. Actual savings therefore depend on the workload and its sharing boundary.

\textbf{Storage and lifecycle cost.} Let $d_{\mathrm{img}}$ be the immutable image already shared by either design, $d_{\mathrm{init}}$ the initialized writable state materialized for one complete runtime, and $d(\Delta_i)$ the subsequent private data written by environment $i$. With capsule metadata cost $d_{\mathrm{meta}}$,
\begin{equation}
\begin{aligned}
    \mathcal{D}_{\mathrm{rep}}(N)
        &=d_{\mathrm{img}}+Nd_{\mathrm{init}}+\sum_{i=1}^{N}d(\Delta_i),\\
    \mathcal{D}_{\mathrm{CUA}}(N)
        &=d_{\mathrm{img}}+d_{\mathrm{init}}+\sum_{i=1}^{N}d(\Delta_i)+d_{\mathrm{meta}}.
\end{aligned}
\label{eq:storage-model}
\end{equation}
This comparison attributes savings to initialized writable state, not to duplicating immutable images. Likewise, lifecycle operations avoid rebuilding the complete runtime but still pay for quiescence, state preparation, rebinding, and readiness checks. Their latency depends on the selected backends and the amount of private state, so the operation-specific measurements in Table~\ref{tab:system_efficiency} complement this accounting. The model describes the expected sources of savings; it does not by itself establish throughput or environment density as concurrency increases.
\begin{table*}[t]
\centering
\caption{
Concurrency scaling measurements at 4 and 16 active environments.
Lower is better for preparation time, batch wall time, RAM, and storage;
higher is better for throughput.
}
\label{tab:concurrency-4-16}
\resizebox{\textwidth}{!}{%
\begin{tabular}{llrrrrrr}
\toprule
Benchmark & Backend & Concurrency &
Env. prep. avg./env (s) &
Batch wall (s) &
Throughput (env/s) &
RAM / env (MiB) &
Storage / env (MiB) \\
\midrule
WebArena & CUA-Sandbox & 4
& 20.360 & 21.340 & 0.188 & 492.9 & 13.00 \\
WebArena & Docker & 4
& 44.600 & 44.800 & 0.089 & 2112.3 & 6757.10 \\
WebArena & CUA-Sandbox & 16
& 44.238 & 49.372 & 0.324 & 436.1 & 13.05 \\
WebArena & Docker & 16
& 120.771 & 121.483 & 0.132 & 2082.6 & 6778.84 \\
\addlinespace
VisualWebArena  & CUA-Sandbox & 4
& 1.434& 5.735 & 0.697 & 15.9 & 0.44 \\
VisualWebArena & Docker & 4
& 6.208 & 24.831 & 0.161 & 514.8 & 4880.04 \\
VisualWebArena  & CUA-Sandbox & 16
& 1.366 & 21.851 & 0.732 & 15.6 & 0.99 \\
VisualWebArena  & Docker & 16
& 4.803 & 76.848 & 0.208 & 349.1 & 4903.34 \\
\addlinespace
OSWorld & CUA-Sandbox & 4
& 5.490 & 8.310 & 0.481 & 851.0 & 28.60 \\
OSWorld & Docker & 4
& 17.870 & 23.950 & 0.167 & 2737.7 & 68.90 \\
OSWorld & CUA-Sandbox & 16
& 7.268 & 10.545 & 1.517 & 836.5 & 29.52 \\
OSWorld & Docker & 16
& 18.742 & 54.293 & 0.295 & 2647.0 & 233.47 \\
\bottomrule
\end{tabular}%
}

\end{table*}

\textbf{Concurrency scaling.}
Table~\ref{tab:concurrency-4-16} evaluates whether the resource and
throughput advantages of CUA-Sandbox persist as execution concurrency
increases from 4 to 16. Across all three workloads, CUA-Sandbox maintains
lower batch latency, higher throughput, and substantially lower
per-environment memory and storage than Docker at both concurrency levels.

The advantage remains substantial at concurrency 16: CUA-Sandbox achieves
2.45$\times$, 3.52$\times$, and 5.14$\times$ higher throughput on
WebArena, VisualWebArena, and OSWorld, respectively. Per-environment memory
is reduced by 4.78$\times$, 22.38$\times$, and 3.16$\times$ on the same
workloads. These results indicate that the efficiency gains are not limited
to the eight-way setting used in the main rollout experiment, but persist
across both lower and higher tested concurrency levels.

\begin{table}[t]
\centering
\caption{
Ablation of copy-on-write (CoW) and runtime sharing for eight logical
environments. Runtime RAM measures the memory occupied by the application
runtime, while Agent/Browser PSS reports the proportional-set size of the
agent and browser processes. Writable storage includes environment-specific
writable state. Lower is better.
}
\label{tab:runtime-sharing-ablation}
\resizebox{\linewidth}{!}{
\begin{tabular}{lcccc}
\toprule
Configuration
& Active runtime containers
& Runtime RAM (MiB)
& Agent/Browser PSS (MiB)
& Writable storage (GiB) \\
\midrule
Official Docker
& 8 Web
& 15,560
& 6,998
& 52.99 \\

CoW + independent runtime
& 8 Web + 8 MySQL
& 13,882
& 7,045
& 24.65 \\

CoW + shared runtime
& 1 Web + 1 MySQL
& \textbf{3,261}
& 7,291
& \textbf{23.19} \\
\bottomrule
\end{tabular}
}
\end{table}
\section{Runtime-sharing ablation.}
\label{sec:ablation33}
Table~\ref{tab:runtime-sharing-ablation} separates the effects of
copy-on-write state management from runtime sharing. Introducing CoW while
retaining independent runtimes reduces writable storage from 52.99 to
24.65~GiB, but only modestly reduces runtime memory from 15,560 to
13,882~MiB. Sharing the initialized runtime further reduces runtime RAM to
3,261~MiB, a 4.26$\times$ reduction relative to the CoW configuration with
independent runtimes and a 4.77$\times$ reduction relative to the official
Docker baseline.

In contrast, Agent/Browser PSS remains approximately unchanged across the
three configurations (6,998--7,291~MiB). This indicates that the memory
reduction does not arise from shrinking the agent or browser workload, but
from eliminating replicated application-runtime processes. Writable storage
changes only slightly when moving from independent to shared runtimes
(24.65 to 23.19~GiB), consistent with CoW accounting for most of the
state-storage reduction while runtime sharing primarily removes replicated
process memory.

\clearpage
\section{Case Studies: State-Scoped Execution}
\label{sec:mechanism-cases}
We illustrate the method in Section~\ref{sec:method} through two task workflows: updating a shopping cart and editing a desktop document. Each follows capsule construction, state-scoped interaction, and reset. The workflows are design illustrations rather than recorded evaluation trajectories.

\begin{figure}[H]
\centering
\begin{minipage}{\linewidth}
\casebox{CaseInk}{\color{white}Case 1: An independent cart over a shared web runtime}{%
\casefield{Scenario.}{Two environments start from the same shopping state. Environment A adds a product to its cart; B should retain its own cart.}
\casefield{Shared / private.}{The initialized web server is shared. Each environment retains its own database branch, session state, cache namespace, and browser profile.}
\vspace{5pt}
\casefield{State sketch.}{A: initial cart $\rightarrow$ updated cart $\rightarrow$ initial cart.\\
B: initial cart throughout, under its own capsule bindings.}}

\noindent\textbf{Phase I: Execute against the selected environment state.}\par\vspace{5pt}
\casebox{ResultOurs}{Step 1: Bind the browser action}{%
\casefield{Action.}{A issues an ``add to cart'' interaction through the original interface.}
\casefield{Binding.}{Resolve A's current generation and lease its database, session, and cache bindings before servicing the request.}}
\casebox{CaseGray}{Step 2: Keep the mutation and observation together}{%
\casefield{State access.}{Resolve the cart update through A's database branch and session. Any resulting cache access or background work must retain A's context.}
\casefield{Expected behavior.}{A sees the added product; B reads its separate state. The evaluator uses A's generation to inspect A's cart.}}

\noindent\textbf{Phase II: Reset one environment without resetting the other.}\par\vspace{5pt}
\casebox{ResultOurs}{Step 3: Stage a clean successor for A}{%
\casefield{Transition.}{Freeze A's route and drain its in-flight work. Prepare a clean database branch, session state, and derived-state namespace as one successor capsule.}
\casefield{Boundary.}{Keep the previous generation published until its successor is ready; B's bindings are not part of this transition.}}
\casebox{CaseGray}{Step 4: Publish bindings and resume interaction}{%
\casefield{Activation.}{Publish A's new generation atomically, reject stale-generation requests, and resume A only after readiness checks.}
\casefield{Expected behavior.}{A returns to the task's initial cart state; B retains its own state. The shared web server remains initialized.}}

\noindent\textbf{Analysis.} The sharing boundary is the request context, not the cart itself. Connection reuse, cache lookup, and asynchronous work must respect that context; an adapter that cannot do so requires a private backend.
\end{minipage}
\caption{\lead{Web case study: request-scoped state separation.} The example follows a cart mutation and an independent reset through shared execution, private state access, and generation publication. Purple headers mark binding and staging; gray headers mark state effects and activation. Expected outcomes illustrate the method's requirements and are not measured execution results.}
\label{fig:web-case}
\end{figure}

\clearpage
\begin{figure}[H]
\centering
\begin{minipage}{\linewidth}
\casebox{CaseInk}{\color{white}Case 2: Document editing with a private state capsule}{%
\casefield{Scenario.}{Two desktop environments open the same initial document. A edits and saves it; B should continue to see its own copy.}
\casefield{Shared / private.}{Compatible runtime components are reusable. Documents, profiles, configuration, displays, and message-bus endpoints remain private to each session.}
\vspace{5pt}
\casefield{State sketch.}{A: initial document $\rightarrow$ saved edit $\rightarrow$ initial document.\\
B: initial document throughout; its session remains separate.}}

\noindent\textbf{Phase I: Bind a desktop interaction to one session.}\par\vspace{5pt}
\casebox{ResultOurs}{Step 1: Activate A's complete session context}{%
\casefield{Binding.}{Attach A's file and profile state, select its display and message-bus endpoints, and establish readiness before accepting input.}
\casefield{Boundary.}{Any process whose mutable state cannot be isolated or safely rebound must remain within a private backend.}}
\casebox{CaseGray}{Step 2: Edit, save, and observe through A's bindings}{%
\casefield{Action.}{A edits the document and saves through the desktop interface.}
\casefield{Expected behavior.}{The save reaches A's file overlay. Screenshots come from A's display, and evaluation reads A's document; B's document is unchanged.}}

\noindent\textbf{Phase II: Reset state before reusing an execution slot.}\par\vspace{5pt}
\casebox{ResultOurs}{Step 3: Quiesce the session and stage its successor}{%
\casefield{Transition.}{Freeze A's route and drain pending input, saves, and background work. Stage clean document and profile state and recreate ephemeral endpoints.}
\casefield{Boundary.}{An open document buffer or file handle must not survive into the successor with stale state; use a private backend where necessary.}}
\casebox{CaseGray}{Step 4: Publish and verify the rebound context}{%
\casefield{Activation.}{Publish only the complete successor bindings and pass readiness checks before routing input or collecting screenshots.}
\casefield{Expected behavior.}{A observes the initial document through the new session. B retains its state, and delayed requests cannot access A's successor generation.}}

\noindent\textbf{Analysis.} The capsule couples document state with the session that exposes it. File isolation alone is insufficient: observations and evaluation must follow the same environment bindings. Components that cannot be safely rebound remain in the coarse private backend described in Section~\ref{sec:state-capsules}.
\end{minipage}
\caption{\lead{Desktop case study: session-scoped state separation.} The workflow follows a document edit through private file and profile state, environment-specific display endpoints, and a coordinated reset. Observations and evaluation follow the same capsule. The private-session boundary also covers components whose in-memory state or open handles cannot be safely rebound.}
\label{fig:desktop-case}
\end{figure}
\clearpage

\end{document}

%% file: ablation-figure.tex
\begingroup
\colorlet{AblFull}{ResultOurs}
\colorlet{AblInk}{ResultOurs!45!black}
\definecolor{AblRemoved}{HTML}{D4D6DD}
\definecolor{AblMuted}{HTML}{666A75}
\begin{tikzpicture}[x=1pt,y=1pt,font=\small]
\path[use as bounding box] (0,-3) rectangle (396,247);
\path[fill=AblRemoved,draw=none] (90.0000,234.0000) rectangle (104.0000,242.0000);
\node[inner sep=0pt,anchor=west,font=\footnotesize] at (110.0000,238.0000) {Corresponding ablation};
\path[fill=AblFull,draw=AblInk,line width=0.25pt] (240.0000,236.0000) rectangle (254.0000,240.0000);
\node[inner sep=0pt,anchor=west,font=\footnotesize] at (260.0000,238.0000) {CUA-Sandbox};
\node[inner sep=0pt,anchor=west] at (0.0000,220.0000) {\textbf{(a) Copy-on-write}};
\node[inner sep=0pt,anchor=west,font=\footnotesize,text=AblMuted] at (151.0000,220.0000) {Without CoW; latency (s)};
\node[inner sep=0pt,anchor=east,font=\footnotesize,text=AblMuted] at (396.0000,220.0000) {Speedup};
\path[fill=AblRemoved,draw=none] (86.0000,196.5000) rectangle (283.1360,205.5000);
\path[fill=AblFull,draw=AblInk,line width=0.25pt] (86.0000,199.0000) rectangle (88.5560,203.0000);
\node[inner sep=0pt,anchor=west] at (0.0000,201.0000) {Create};
\node[inner sep=0pt,anchor=west,text=AblInk,font=\footnotesize] at (90.5560,209.0000) {0.071};
\node[inner sep=0pt,anchor=west,text=AblMuted,font=\footnotesize] at (287.1360,201.0000) {5.476};
\node[inner sep=0pt,anchor=east,text=AblInk,font=\footnotesize] at (396.0000,201.0000) {\textbf{77.1$\times$}};
\path[fill=AblRemoved,draw=none] (86.0000,177.5000) rectangle (255.0200,186.5000);
\path[fill=AblFull,draw=AblInk,line width=0.25pt] (86.0000,180.0000) rectangle (88.9520,184.0000);
\node[inner sep=0pt,anchor=west] at (0.0000,182.0000) {Checkpoint};
\node[inner sep=0pt,anchor=west,text=AblInk,font=\footnotesize] at (90.9520,190.0000) {0.082};
\node[inner sep=0pt,anchor=west,text=AblMuted,font=\footnotesize] at (259.0200,182.0000) {4.695};
\node[inner sep=0pt,anchor=east,text=AblInk,font=\footnotesize] at (396.0000,182.0000) {\textbf{57.3$\times$}};
\path[fill=AblRemoved,draw=none] (86.0000,158.5000) rectangle (257.7920,167.5000);
\path[fill=AblFull,draw=AblInk,line width=0.25pt] (86.0000,161.0000) rectangle (87.9440,165.0000);
\node[inner sep=0pt,anchor=west] at (0.0000,163.0000) {Fork};
\node[inner sep=0pt,anchor=west,text=AblInk,font=\footnotesize] at (89.9440,171.0000) {0.054};
\node[inner sep=0pt,anchor=west,text=AblMuted,font=\footnotesize] at (261.7920,163.0000) {4.772};
\node[inner sep=0pt,anchor=east,text=AblInk,font=\footnotesize] at (396.0000,163.0000) {\textbf{88.4$\times$}};
\node[inner sep=0pt,anchor=west,font=\footnotesize,text=AblMuted] at (86.0000,148.0000) {Storage delta: \textcolor{AblInk}{0.11 MiB} (full) / 3.67 GiB (without CoW)};
\node[inner sep=0pt,anchor=west] at (0.0000,127.0000) {\textbf{(b) Non-DB state}};
\node[inner sep=0pt,anchor=west,font=\footnotesize,text=AblMuted] at (151.0000,127.0000) {Without non-DB state; latency (s)};
\node[inner sep=0pt,anchor=east,font=\footnotesize,text=AblMuted] at (396.0000,127.0000) {Added time};
\path[fill=AblRemoved,draw=none] (86.0000,102.5000) rectangle (128.4600,111.5000);
\path[fill=AblFull,draw=AblInk,line width=0.25pt] (86.0000,105.0000) rectangle (174.8800,109.0000);
\node[inner sep=0pt,anchor=west] at (0.0000,107.0000) {Prepare};
\node[inner sep=0pt,anchor=west,text=AblInk,font=\footnotesize] at (176.8800,115.0000) {1.616};
\node[inner sep=0pt,anchor=west,text=AblMuted,font=\footnotesize] at (132.4600,99.0000) {0.772};
\node[inner sep=0pt,anchor=east,text=AblInk,font=\footnotesize] at (396.0000,107.0000) {+0.844 s};
\path[fill=AblRemoved,draw=none] (86.0000,81.5000) rectangle (190.9950,90.5000);
\path[fill=AblFull,draw=AblInk,line width=0.25pt] (86.0000,84.0000) rectangle (250.7250,88.0000);
\node[inner sep=0pt,anchor=west] at (0.0000,86.0000) {Reset};
\node[inner sep=0pt,anchor=west,text=AblInk,font=\footnotesize] at (252.7250,94.0000) {2.995};
\node[inner sep=0pt,anchor=west,text=AblMuted,font=\footnotesize] at (194.9950,78.0000) {1.909};
\node[inner sep=0pt,anchor=east,text=AblInk,font=\footnotesize] at (396.0000,86.0000) {+1.086 s};
\node[inner sep=0pt,anchor=west] at (0.0000,65.0000) {\textbf{(c) Publication}};
\node[inner sep=0pt,anchor=west,font=\footnotesize,text=AblMuted] at (151.0000,65.0000) {Direct publication; events / 100 trials};
\node[inner sep=0pt,anchor=east,font=\footnotesize,text=AblMuted] at (396.0000,65.0000) {};
\path[fill=AblRemoved,draw=none] (86.0000,39.5000) rectangle (286.0000,48.5000);
\draw[AblInk,line width=0.9pt] (86.0000,41.5000) -- (86.0000,46.5000);
\node[inner sep=0pt,anchor=west] at (0.0000,44.0000) {Mixed generations};
\node[inner sep=0pt,anchor=west,text=AblInk,font=\footnotesize] at (88.0000,52.0000) {0/100};
\node[inner sep=0pt,anchor=west,text=AblMuted,font=\footnotesize] at (290.0000,44.0000) {100/100};
\node[inner sep=0pt,anchor=east,text=AblInk,font=\footnotesize] at (396.0000,44.0000) {100 fewer};
\path[fill=AblRemoved,draw=none] (86.0000,18.5000) rectangle (286.0000,27.5000);
\draw[AblInk,line width=0.9pt] (86.0000,20.5000) -- (86.0000,25.5000);
\node[inner sep=0pt,anchor=west] at (0.0000,23.0000) {In-flight publication};
\node[inner sep=0pt,anchor=west,text=AblInk,font=\footnotesize] at (88.0000,31.0000) {0/100};
\node[inner sep=0pt,anchor=west,text=AblMuted,font=\footnotesize] at (290.0000,23.0000) {100/100};
\node[inner sep=0pt,anchor=east,text=AblInk,font=\footnotesize] at (396.0000,23.0000) {100 fewer};
\node[inner sep=0pt,anchor=west,font=\footnotesize,text=AblMuted] at (86.0000,3.0000) {Frozen requests rejected: 100/100 (full) / N/A (direct)};
\end{tikzpicture}
\endgroup